# ANTLER: Bayesian Nonlinear Tensor Learning and Modeler for Unstructured, Varying-Size Point Cloud Data

Michael Biehler, *Member, IEEE*, Hao Yan, and Jianjun Shi

*Abstract* — **Unstructured point clouds with varying sizes are increasingly acquired in a variety of environments through laser triangulation or Light Detection and Ranging (LiDAR). Predicting a scalar response based on unstructured point clouds is a common problem that arises in a wide variety of applications. The current literature relies on several pre-processing steps such as structured subsampling and feature extraction to analyze the point cloud data. Those techniques lead to quantization artifacts and do not consider the relationship between the regression response and the point cloud during pre-processing. Therefore, we propose a general and holistic "Bayesian Nonlinear Tensor Learning and Modeler" (ANTLER) to model the relationship of unstructured, varying-size point cloud data with a scalar or multivariate response. The proposed ANTLER simultaneously optimizes a nonlinear tensor dimensionality reduction and a nonlinear regression model with a 3D point cloud input and a scalar or multivariate response. ANTLER has the ability to consider the complex data representation, high-dimensionality, and inconsistent size of the 3D point cloud data.**

*Note to practitioners* — **This paper is motivated by a real-world case study concerning the prediction of the transmission error and eccentricity based on unstructured point clouds of varying size in gear manufacturing. In the current state-of-the-art method, those characteristics can only be obtained via expensive and time-consuming Finite Element Analysis (FEA) or testbenches. The proposed ANTLER framework can directly link the measurement point clouds with a scalar response and serves as a guiding example for the immense potential of the ANTLER.**

## I. INTRODUCTION

THREE-DIMENSIONAL (3D) point cloud acquisition devices, such as laser scanners for surface modeling and geometric reconstruction, have created vast amounts of unstructured 3D point clouds. Previous work has mainly addressed the segmentation [1] and classification [2] of objects based on point clouds. However, the nonlinear modeling of high-dimensional point cloud data in the engineering domain with limited sample sizes has received little attention. To this end, it is necessary to propose a holistic one-step framework that can combine extraction of nonlinear features and estimation of a nonlinear relationship between unstructured point clouds and a scalar response. Such modeling problems based on point clouds have essential applications, for example, in the manufacturing domain. In additive manufacturing (AM), the prediction of part distortion [3] based on the 3D layer shapes for a given process setting relies heavily on time-consuming FEA, which is infeasible for in-process real-time functional quantification and verification. A holistic machine learning approach to directly predict the response (e.g., part distribution) is needed for real-time monitoring. The application that motivated this research is the prediction of the transmission error of gears based on in-line 3D measurement data (Fig. 1).

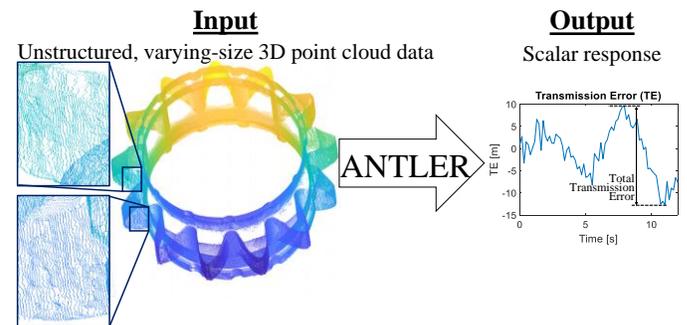

**Fig. 1.** Illustration of the unstructured point cloud from in-line 3D measurement and transmission error

The point clouds presented in the micro gear case study (Section V) exhibit a highly unstructured point distribution (see magnifications in Fig. 1) due to the acquisition by optical metrology. This also results in a varying number of measurement points across samples. An essential key quality indicator which is mainly caused by manufacturing inaccuracies is the transmission error. In the current state-of-the-art, the modeling of the relationship between the point clouds and the transmission error depends on expensive experiments or time-consuming FEA simulations, which cannot be applied to in-process quality improvement

M. Biehler is with the H. Milton Stewart School of Industrial and Systems Engineering (ISyE), Georgia Institute of Technology, Atlanta, GA 30332 USA (e-mail: michael.biehler@gatech.edu).

H. Yan is with the School of Computing, Informatics & Decision Systems Engineering, Arizona State University, Tempe, AZ 85281 USA (e-mail: haoyan@asu.edu).

J. Shi is with the H. Milton Stewart School of Industrial and Systems Engineering (ISyE), Georgia Institute of Technology, Atlanta, GA 30332 USA (e-mail: jianjun.shi@isye.gatech.edu). Dr. Shi is the corresponding author.

methodologies. Therefore, a data-driven method is needed to directly relate the point cloud with the transmission error.

The main goal of this article is to propose a novel nonlinear regression method to quantify the relationship between unstructured, varying-size point clouds and a scalar or multivariate, vector response. However, there are significant challenges in achieving this objective due to the complex data format of the unstructured 3D point clouds:

- *Unstructured*: Typically, there is no prior structure of the set of 3D coordinates for complex shapes. In particular, their data structure is not topologically aware, and the spatial neighboring relationship among points is unknown, which leads to significant modeling challenges for nonlinear point cloud regression.

- *High dimensionality*: Point clouds are spatially dense, containing a large number of measurement points (e.g., millions) in each sample, which poses significant computational challenges.

- *Limited sample size*: On the contrary, the number of samples in the engineer applications is typically very limited and much smaller than the number of dimensions.

- *The different number of points per sample*: The number of measurement points varies per sample, making the adaptation of machine learning techniques challenging because most machine learning models require designed input training data with a regular structure and fixed dimensionality.

To tackle those challenges, existing research has mainly focused on combining dimensionality reduction and regression models for structured point cloud data. The structure allows the efficient representation of the point clouds as tensors [4]. However, these methods cannot be applied directly to unstructured point cloud data without subsampling or interpolation, which typically only preserves global spatial information while losing local spatial information. In particular, detail-level spatial information such as local features or anomalies are lost. For a more detailed review of the tensor-based structured point cloud methods, readers are referred to Section II. A.

Recently, several deep learning-based methods have been proposed for unstructured point cloud data [1, 5]. These methods have the ability to learn the low-dimensional representation directly from the raw point cloud data but require a large number of training samples, which may not be feasible in engineering applications. For a more detailed review of the deep-learning-based point cloud data regression and classification, please refer to Section II. C.

To address these aforementioned challenges, we propose a novel method, namely "Bayesian Nonlinear Tensor Learning and ModelER" (ANTLER), to model the unstructured, varying-size point cloud data by adopting a Variational Autoencoder with a novel loss function to simultaneously optimize the nonlinear dimensionality reduction via nonlinear tensor decomposition and nonlinear regression while preserving the computational efficiency. The unified approach can learn the low-dimensional data representation most correlated to the response data.

The remainder of the article is organized as follows. Section II gives a brief literature review on point cloud modeling. Then the proposed ANTLER framework for an unstructured, varying-size 3D point cloud input and a scalar response is introduced in Section III. Section IV validates the proposed methodology by using simulated data with two different types of unstructured point clouds. Furthermore, the performance of the proposed method is compared with existing benchmark methods in terms of estimation accuracy and computational time. In Section V, we conduct a real-world case study for predicting the functional response in gear manufacturing applications. Finally, we conclude the article with a short discussion and an outline of future work in Section VI.

## II. LITERATURE REVIEW

In this section, we will review three major categories of methods for point cloud modeling problems, including tensor regression, feature extraction-based, and deep learning methods. We note, to the best of our knowledge, no existing research has directly addressed regression problems with unstructured, varying-size 3D point clouds as inputs and scalar or multivariate responses. A wide variety of machine learning methods, such as traditional regression methods, are not applicable to this problem since vectorized unstructured point clouds are not permutation invariant. Therefore, their spatial relationship cannot be preserved because the order of measured locations in the vectorized format may change with every sample. Consequently, the remainder of this section only reviews methods that can be extended to model the unstructured point clouds. Most of the latter methods cannot implicitly handle input point clouds of varying size.

### A. Tensor Regression Techniques for Structured Point Cloud Data

While classical regression techniques treat covariates as vectors, the rise of sensing technology and high-dimensional (heterogeneous) data has led to covariates of more complex forms such as multidimensional arrays, also known as tensors. However, these methods require the efficient representation of the data as a tensor. Therefore, in the domain of point cloud data, the data type of structured point cloud has been the focus of attention. In structured point clouds, the data lies in a predefined grid and can be efficiently represented as a tensor. Along this direction, Yan et al. [4] proposed a regularized tensor regression for structured point cloud data analysis, which links variational patterns of point clouds (response) and to the process variables (inputs). Gahrooei et al. [6] has proposed a multiple tensor-on-tensor (MTOT) regression framework to model the behavior of a system as a function of heterogeneous sets of data, such as scalars, waveform signals, images, or structured point clouds. Wang et al. [7] proposed an augmented tensor regression model for settings with missing data. By integrating tensor regression and tensor completion, they can handle scenarios where measurements of the 3D point clouds are incomplete.

However, these methods have the following limitations: 1) These methods require point clouds with structured grid measurements and a fixed number of measurements. 2) These approaches all rely on a linear model formulation using Tucker decomposition, which is not suitable for many engineering applications with nonlinear relationships.

### B. Feature Extraction-based Approach for Point Cloud Data

The current literature on point cloud modeling mainly utilizes a two-step approach, which first extracts useful features and then uses them as an input to an off-the-shelf predictive model. A wide range of features such as dissimilarity metrics [8], principal components [9], or geometric features [10] have been proposed to this end. Furthermore, several methods are based on tensor voting [2, 11, 12]. For example, Du et al. (2021) [2] proposed a tensor voting-based surface anomaly classification approach based on 3D point cloud data. Features obtained from the point clouds via saliency derivation and sharp point selection are used to train a sparse multiclass surface anomaly classifier.

### C. Deep Learning Methods for Point Cloud Data

Recently, deep learning methods have been developed for classification and object segmentation of the point cloud data due to a large amount of public 3D point cloud datasets. Most prominently, [1] proposed a deep neural network (PointNet) that uses point cloud inputs for various 3D recognition tasks such as object classification, part segmentation, and semantic segmentation. Early works utilize the distribution of the points to aid the classification of 3D objects [13]. Another direction of deep learning research utilizes graph neural networks for object detection [5], classification [14], and segmentation [15]. In particular, [5] encodes the point cloud efficiently in a fixed radius near-neighbors graph and then utilizes a graph neural network to predict the category and shape of objects. Similarly, [14] proposed a graph attention-based point neural network to learn shape representations from point clouds. [15] utilized spectral graph theory to represent point cloud features as signals on a graph and applied a graph convolutional neural network on the graph Laplacian matrix.

The main drawback of those deep learning methods is the difficulty in capturing sparse local features for discriminative tasks. Therefore, they are not suitable for engineering applications with small engineering tolerance and consequent high precision requirements for discriminative models. Additionally, those techniques can lead to computational challenges due to the high dimensionality of the point clouds. The large number of model parameters to be estimated can also require a very large number of training samples. Therefore, it may lead to overfitting or inaccurate models, especially when the number of training samples is small.

In summary, there is very limited work on regressions with unstructured point cloud input and scalar output, which can address the aforementioned challenges. This paper fills these research gaps and proposes a method for the prediction of scalar or vector responses based on an unstructured, varying-size point cloud input.

## III. ANTLER METHODOLOGY

In this section, we introduce the ANTLER framework as an approach to model a scalar response as a function of an unstructured, varying-size 3D point cloud input.

We assume that a set of unstructured 3D input point clouds of size $N$ is available, denoted as $\mathbf{X}_i \in \mathbb{R}^{3 \times M_i}$ of varying-size $M_i$ and a response vector $Y_i \in \mathbb{R}^p$, where $p$ is the number of outputs, and $i$ is the sample index. The ANTLER framework first represents the unstructured point cloud $\mathbf{X}$ as a binary tensor $\mathcal{B}$ utilizing a voxel-based representation (Section III.A). Afterwards, an efficient sampling strategy (Section III.B) is utilized to obtain a balanced sampling tensor $\mathcal{D}$.

Subsequently, a Streaming Nonlinear Bayesian Tensor Decomposition (SNBTD) (Section III.C) is utilized to obtain a low dimensional embedding $\mathcal{U}$ of the binary tensor $\mathcal{B}$. In particular, random Fourier features are used to obtain a sparse spectrum Gaussian Process (GP) decomposition model, where the joint probability model of $K$ latent embedding matrices $\mathcal{U} = \{\mathbf{U}^1, \ldots, \mathbf{U}^K\}$ becomes

$$p(\mathcal{U}, \mathbf{S}, \mathbf{w}, \mathcal{D}) = \prod_{k=1}^{K} \prod_{j=1}^{d_k} \prod_{t=1}^{r_k} \mathcal{N}(u_{jt}^k | 0,1)$$
$$\cdot \prod_{m=1}^{M} \prod_{j=1}^{R} \mathcal{N}(s_{mj} | 0,1) \mathcal{N}(\mathbf{w} | 0, M^{-1}\mathbf{I})$$
$$\cdot \prod_{i \in D} \Phi\big((2a_{\mathbf{i}_n} - 1)\mathbf{w}^T \phi(\mathbf{x_i})\big), \quad (1)$$

In Section III.D, we link the low dimensional embedding $\mathcal{U}$ and the scalar response $Y$ via a nonlinear regression method. In particular, $Y = g_{\theta_r}(\mathcal{U}) + E$, where $\theta_r$ are the model parameters to be estimated, and $E$ is the error tensor.

Finally, we utilize a Variational Autoencoder (VAE) with a novel ANTLER loss to simultaneously learn the low dimensional embedding $\mathcal{U}$ and find the nonlinear regression relationship between $\mathcal{U}$ and the response $Y$. Interested readers are referred to Doersch [16] and the references therein for a detailed review of the VAE framework.

The unified framework for the simultaneous optimization of the decomposition of the binary tensor $\mathcal{B}$ into the low dimensional embedding $\mathcal{U}$ and regression function (i.e., $Y = g(\mathcal{U})$) enables the manifold learning (Section III.E) to capture the essential information necessary for the discriminative task and optimize the proposed representation. This procedure is summarized in Fig. 2.

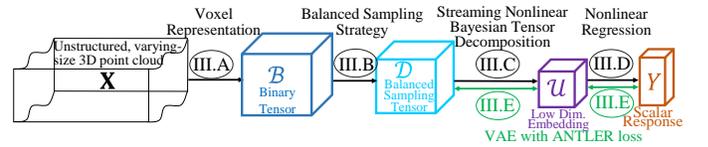

**Fig. 2**: Proposed ANTLER Framework

The main underlying assumption for the ANTLER framework is the manifold hypothesis, which states that high-dimensional observations typically lie in a much lower-dimensional, not necessarily linear, manifold.

### A. Binary 3D Tensor Representation of Point Cloud Data

Several approaches to represent surfaces of 3D objects, such as triangular meshes or graphs, have been proposed in the literature. However, there are three main requirements regarding the representation of unstructured, varying-size point

clouds:

1. Representation should consider the spatial correlation structure in unstructured point clouds.
2. Representations should respect the permutation invariance of the input point clouds [1].
3. Representations should have fixed dimensions instead of the original varying dimensions of the point cloud data.

Therefore, we take advantage of the volumetric voxel representation of point clouds [17]. This approach allows the transformation of unstructured point clouds into regular voxel grids. Each voxel contains a Boolean occupancy status (i.e., occupied by measurement point or unoccupied).

In our setting, the original point cloud $\mathbf{X}_i$ is represented as a binary 3D tensor $\mathcal{B}_i$ utilizing a 3D grid sufficiently larger than the measurement object to ensure all measurement data is contained inside the grid (Fig. 3a). On this cube, a 3D structured grid with a prespecified grid width and height is defined. Then each grid will be assigned with binary values: If there are measurement points in $\mathcal{B}_i$, the value in that particular grid region $\mathcal{B}_i$ is assigned as 1. If there is no measurement point in a region, the corresponding region will have entry 0 (Fig.3b and 3c).

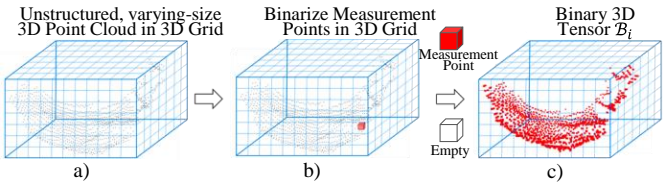

**Fig.3:** Process to obtain structured fixed-size binary 3D tensor from unstructured varying-size point cloud

The recommended strategy for selecting the grid size is as follows: The grid size is determined by starting from a prespecified (small) grid size (e.g., $100 \times 100 \times 100$). Then, the resolution of the grids is increased until the number of non-zero elements in the binary tensor is equal to the number of measurement points in each training sample. We note that due to the subsequent balanced sampling (Algorithm 1) the grid size of each sample is not required to be the same for each sample. The main reasons for selecting the voxel-based representation are two-fold: Firstly, they allow the efficient estimation of occupied and unoccupied space from a wide range of measurements, even when the measurement conditions (e.g., acquisition angle, lighting conditions) are changing. Secondly, they can be stored and manipulated with efficient data structures such as tensors.

We note that the voxel-based representation of point clouds has already been proposed almost 40 years ago [18], but has not gained enough popularity due to the inefficiency of such representations. This is due to the extreme data sparsity and the computational cost, given that many tensor voxel entries would be empty, and the dimensionality of the voxel representation is typically high. To this end, in Section III.B, we propose how an efficient sampling strategy can learn the representation from very sparse binary voxel tensors.

## B. Efficient Balanced Sampling Method for Extremely Sparse Tensors

We first would like to point out that the original SNBTD algorithm suffers from data imbalance issues, given that the number of non-zero entries is extremely sparse. This would make the learning algorithm extremely biased towards the zero entries by reconstructing all entries as zero. For example, in the case study presented in Section V, we can achieve an average accuracy of 99.7% by only reconstructing zero entries since the binary only contains 0.3% non-zero entries on average. To address this extreme sparsity issue, we utilize a balanced sampling algorithm for the entries of a binary tensor sample $\mathcal{B}_i$ and obtain a 4D array $\mathcal{D}_i$ with a balanced number of zero and non-zero entries. The key idea of the sampling strategy is as follows: The k-nearest neighborhood sampling imitates the shape of the measurement objects by learning the surface boundary in the vicinity of the measurements. If domain knowledge is available about surface boundaries in the vicinity of the measurement points, it can be incorporated into the selection of meaningful entries for training. The realization of this sampling strategy is shown in Algorithm 1.

---

**Algorithm 1:** Binary Tensor to 4D array Sampling Strategy

**Input:**
  Binary Tensor representation $\mathcal{B}_i$ of unstructured-varying size point clouds $\mathbf{X}_i$ obtained via strategy from Section III.A

**Result:**
  $\mathcal{D}_i$: Balanced 4D sampling array

**Algorithm:**
Initialize: $M_r = 2 \cdot \max_{i=1,\dots,n} M_i$, $\mathcal{D}_i \in \mathbb{R}^{4 \times M_r}$, $j = 1, \dots, R \times M_i$, $c = 0$
for $j = 1, \dots, R \times M_i$
  if $b_j = 1$
    $\mathcal{D}_{i,c,:} = (x_j, y_j, z_j, b_j), c = c + 1$
  else if
    $j_z = \{j | b_j = 0\}$
    Sample $M_i$ elements of $j_z$ in the $k = 1$ nearest neighborhood
    $\mathcal{D}_{i,c,:} = (x_{j_z}, y_{j_z}, z_{j_z}, b_{j_z}), c = c + 1$
    if $2 \cdot M_i < M_r$
      Randomly sample $M_r - 2 \cdot M_i$ elements of $j_z$
      $\mathcal{D}_{i,c,:} = (x_{j_z}, y_{j_z}, z_{j_z}, b_{j_z}), c = c + 1$
end if $c = M_r$

---

This algorithm enables the estimation of meaningful tensor decomposition and significantly reduces the computational effort of evaluating the voluminous binary tensor representation. This is a remedy for the drawbacks of the volumetric and sparse point cloud voxel representation. We note that the input point clouds have a varying number of points $M_i$, which will lead to a varying number of non-zero entries in the binary tensor $\mathcal{B}_i$. However, to utilize the VAE based framework introduced in Section III.D, we define a conservative upper bound of $M_r$ elements to ensure a fixed input dimension to the VAE. Consequently, when all $M_i$ non-zero entries of the binary tensor $\mathcal{B}_i$ are sampled, at most $M_r/2$ non-zero elements will be contained in the binary sampling tensor $\mathcal{D}_i$.

## C. Streaming Nonlinear Bayesian Tensor Decomposition (SNBTD) via Stream Patch Processing

To take advantage of the superior performance of Bayesian

Tensor decompositions on small datasets, we will regularize the VAE embeddings in Section III.D with the low dimensional embeddings $\mathcal{U}_{SNBTD}$. Those embeddings are obtained by the SNBTD method, which decomposes the binary sampling tensor $\mathcal{D}_i$ that was obtained from the binary point cloud representation $\mathcal{B}_i$ of the unstructured, varying-size point cloud samples $\mathbf{X}_i$ (Algorithm 1). To make the SNBTD method applicable to the proposed ANTLER framework, several aspects need to be modified. The binary tensors in the proposed framework are extremely sparse (e.g., 0.3% non-zero entries for micro gear dataset (Section V) and also very high dimensional (e.g., $1000 \times 1000 \times 1000 = 1$ Billion entries for micro gear dataset). The SNBTD method, on the other hand, is essentially a sparse spectrum approximation of the Gaussian process decomposition model. This sparse spectrum approximation relies on a small set of pseudo inputs, which have a Gaussian distribution prior. For extremely sparse input tensors, this will lead to a bias towards zero, so only zero entries would be reconstructed. To this end, we proposed an efficient sampling strategy in Section III.B (Algorithm 1). This allows us to obtain a balanced dataset representing the 3D shape of the measured objects.

Another challenge is the high dimensionality of those tensors: to efficiently precompute the low dimensional embeddings $\mathcal{U}_{SNBTD}$ which serves as an input to the VAE with ANTLER loss, we adopt the streaming patch processing for the SNBTD. This method allows the processing of observed tensor entries in small patches by updating the posterior of latent embeddings $\mathcal{U}_{SNBTD}$, the weights w and the frequencies $\mathbf{S}$ based on each patch of the sampling tensor $\mathcal{D}_t$, without using the previously accessed batches $\{\mathcal{D}_1, \ldots, \mathcal{D}_{t-1}\}$. The process of streaming patch processing is visualized in Fig. 4.

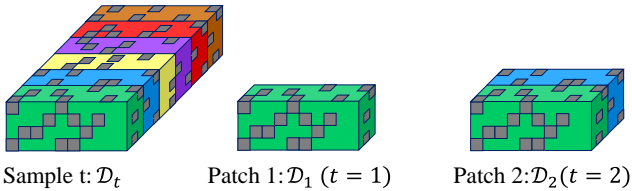

Sample t: $\mathcal{D}_t$    Patch 1: $\mathcal{D}_1$ ($t=1$)    Patch 2: $\mathcal{D}_2$ ($t=2$)

**Fig. 4**: Streaming Patch Processing

As an output of Algorithm 1 (Section III.B), the unstructured point clouds of varying size $\mathbf{X}_i$, which had been represented as a binary tensor $\mathcal{B}_i$, are finally represented as a 4D vector $\mathcal{D}_i$ ($\mathbf{i}, \mathbf{b}_i$), where $\mathbf{i}$ denotes the x,y,z-coordinates of the binary tensor $\mathcal{B}_i$ and the tensor entry value is denoted by $b_i$ (i.e., 0 or 1). For patch processing, the factorized posterior of the binary tensor

$$q(\mathcal{U}, \mathbf{S}, \mathbf{w}) = \prod_{k=1}^{K} \prod_{j=1}^{d_k} \prod_{t=1}^{r_k} q(u_{jt}^k) \prod_{m=1}^{M} \prod_{j=1}^{R} q(s_{mj}) \cdot q(\mathbf{w}) \quad (2)$$

is initialized with the prior, where each $q(u_{jt}^k) = \mathcal{N}(u_{jt}^k | \mu_{jt}^k, v_{jt}^k)$, $q(s_{mj}) = \mathcal{N}(s_{mj} | \alpha_{mj}, \rho_{mj})$ and $q(\mathbf{w}) = \mathcal{N}(\mathbf{w} | \boldsymbol{\eta}, \boldsymbol{\Sigma})$. If some engineering knowledge about the shape of the objects is available beforehand, a more informative prior distribution can be specified. Then the new arriving batch $\mathcal{D}_t$ is

utilized to construct a blending distribution

$$p_b(\theta_{p_b}) = p_b\left(\theta_{p_b \setminus s_{mj}}\right) p_b\left(s_{mj} | \theta_{p_b \setminus s_{mj}}\right), \quad (3)$$

where $\theta_{p_b} = \{\mathcal{U}, \mathbf{S}, \mathbf{w}\}$, $\theta_{p_b \setminus s_{mj}} = \theta_{p_b} \setminus \{s_{mj}\}$ and

$p_b\left(s_{mj} | \theta_{p_b \setminus s_{mj}}\right) \propto q(s_{mj}) \prod_{i_n \in B_t} \Phi((2b_{i_n} - 1) \mathbf{w}^T \boldsymbol{\phi}(\mathbf{x}_i))$.

Finally, $\{q(s_{mj})\}$, $\{q(u_{jt}^k)\}$ and $q(\mathbf{w})$ is updated in parallel with conditional moment matching coupled with Gaussian-Hermite quadrature and Taylor approximation to ensure computational tractability [19].

## D. Nonlinear Regression Approach

The goal of the proposed method, contrary to an abundance of previous research in the field of tensor regressions, is to model the relationship between the low dimensional embedding $\mathcal{U}$ learned in Section III.C and the response $Y$ with the following nonlinear form:

$$Y = g_{\theta_r}(\mathcal{U}) + E, \quad (4)$$

where $\theta_r$ are the model parameters of a nonlinear function $g(\cdot)$ to be estimated, and $E$ is the error tensor. The proposed framework allows the integration of various off-the-shelf nonlinear regression methods such as random forests or deep neural networks. An appropriate method can be chosen based on cross-validation. This overcomes the extreme shortcoming of previous tensor regression methods [4, 6] that rely both on linear decomposition and linear regression techniques, deeming them impractical in many real-world engineering applications, which exhibit complex nonlinear structures.

For the remainder of this paper, without loss of generality, we will utilize Deep Neural Networks (DNN) to perform the nonlinear regression task. This is due to the high expressivity of properly trained DNN's and the implicit self-regularization in DNN's derived by Martin et al. (2018) [20]. Even though the dimensionality of the binary tensor was significantly reduced due to Algorithm 1, the dimension of the low dimensional embeddings $\mathcal{U}$ is still relatively high, which considering the small sample size in many engineering applications may lead to overfitting and the curse of dimensionality. Therefore, we recommend utilizing regularization strategies in the estimation of $\theta_r$ to alleviate representational difficulties and bad generalization behavior. In particular, DNN training exhibits a Tikhonov-like form of self-regularization, which can be enhanced by explicit forms of regularization such as dropout and weight norm constraints. However, the most important tuning parameter during DNN optimization is the batch size: larger batch sizes lead to less-well implicitly regularized models [20]. Following these theoretical results, we recommend a batch size of 1 to achieve the maximal self-regularization in training with small sample sizes. The optimization of the network architecture is conducted based on common techniques such as grid search based on cross validation.

## E. Unified ANTLER Framework for Efficient Inference

In this section, we propose an efficient framework to simultaneously optimize the components of the ANTLER framework and achieve a computationally tractable approximation of the nonlinear tensor decomposition introduced in Section III.C and the nonlinear regression method in Section III.D.

The SNBTD has a time complexity of order $\mathcal{O}(N(\sum_k d_k r_k + MR + 4M^2))$ [19] which is prohibitive for an online setting. Typical values in the proposed setting are a sample size of $N$=100 with $d_k = 1,000,000$ $(k = 1,2,3)$ points in each sample, a dimension of each embedding vector $r_k$ in the range 3-8, $R = \sum_k r_k$ and $M = 128$ independent frequencies. To alleviate the computational burden of the SNBTD, but still exploit its immense nonlinear tensor decomposition capability and simultaneously update the nonlinear tensor decomposition as well as the nonlinear regression, we propose the following novel ANTLER loss function:

$$\mathcal{L}_{\phi,\theta,\theta_r}(\mathcal{D}) = \sum_{\mathcal{D}_i \in \mathcal{D}} \log \frac{1}{S} \sum_{j=1}^{S} \frac{p_\theta(\mathcal{D}_i | \mathbf{z}_{(j)})}{q_\phi(\mathbf{z}_{(j)} | \mathcal{D}_i)}$$
$$+ \lambda_1 D_{KL}\left(q_\phi(\mathbf{z} | \mathcal{D}_i) \| p(\mathbf{z})\right) + \lambda_2 \|\mu_{\mathbf{z}} - \mu_{U_{SNBTD}}\|_2^2$$
$$+ \lambda_3 \|g_{\theta_r}(\mu_{\mathbf{z}}) - Y\|_2^2 , \quad (5)$$

where $\lambda_1, \lambda_2, \lambda_3$ are tuning parameters, $\mu_{\mathbf{z}}$ denotes the mean of $\mathbf{z}$, $\mu_{U_{SNBTD}}$ denote the mean of the low dimensional embedding $\mathcal{U}$ obtained via the SNBTD method and the reparameterization trick is applied as $\mathbf{z} = g(\epsilon, \phi, \mathcal{D}_i)$. The terms in the loss function have the following interpretations:

The *first two terms* represent the original ELBO loss function from the classical VAE. The first term denotes the expected error in reconstructing the binary data tensor $\mathcal{D}$ from z. The second term denotes the Kullback-Leibler (KL) divergence, which ensures that auxiliary distribution q(z) is close in terms of distribution to p(z).

The *third term* explicitly regularizes the solution of VAE to approximate the mean of the low dimensional embedding $\mathcal{U}$ obtained via SNBTD, $\mu_{U_{SNBTD}}$. The additional SNBTD loss alleviates the drawback of VAEs, that priors are often highly redundant due to i.i.d. assumptions on internal parameters [21]. In contrast to vanilla VAEs, this allows the latent embeddings to model the full range of statistical properties of the input data. By enforcing the SNBTD properties, we promote diversity in the latent variables, which is difficult to achieve in the generative framework since VAEs typically learn "incorrect" latent representations [22]. This is important since the salient statistical properties of the low dimensional embedding will be used for discrimination. Embeddings need to span the space of possible data and represent diverse characteristics, which is important for discriminative tasks. However, the SNBTD method is not applied directly since the posterior distribution does not have a closed-form expression. The posterior optimization must rely on variational inference. This is computationally expensive, especially when the number of predictors is large. Therefore, it is necessary to find a more computationally tractable way to represent the SNBTD low-dimensional embeddings. Since a two-layer perceptron can approximate any (nonlinear) function on a bounded region [23, 24], a VAE can be used to approximate the posterior distribution of the SNBTD without the need of the expensive Bayesian inference.

Therefore, we can integrate the advantages of both deep VAEs and Bayesian methods. We note that for $\lambda_2 = 0$, this SNBTD

property of the VAE disappears. However, our experimental results on the synthetic and gear datasets showed that including this term achieves significantly higher prediction performance. The *fourth term* models the nonlinear relationship between the mean $\mu_{\mathbf{z}}$ of the low dimensional embeddings $\mathbf{z}$ and the response $Y$. By integrating this directly into the VAE, we can simultaneously estimate the nonlinear tensor decomposition and nonlinear regression in a one-step approach to alleviate the suboptimality of a two-step approach. Contrary to two-step approaches, this one-step approach enables the recognition of variation patterns that are correlated with the binary input tensor as verified in the simulation studies.

Consequently, the **ANTLER algorithm** is given as follows.

---

**ANTLER:** B<u>A</u>yesian <u>N</u>onlinear <u>T</u>ensor <u>L</u>earning and Mod<u>E</u>l<u>ER</u> for Unstructured, Varying-size Point Cloud Data

**Input:**
    $\mathbf{X} \in \mathbb{R}^{3 \times M_i \times n}, Y \in \mathbb{R}^{p \times n}$: Dataset
    $q_\phi(\mathbf{z} | \mathcal{D})$: Inference model
    $p_\theta(\mathcal{D} | \mathbf{z})$: Generative model
    $g_{\theta_r}(\mathbf{z})$: Nonlinear regression model
    $\lambda_1, \lambda_2, \lambda_3$: Tuning parameters
**Result:**
    $\phi, \theta, \theta_r$: Learned parameters
**Part 1: Binary Tensor Representation**
    Represent the unstructured varying-size point cloud samples
    $\mathbf{X}_i \in \mathbb{R}^{3 \times M_i}$ as binary tensors $\mathcal{B}_i \in \mathbb{R}^{d_{x,i} \times d_{y,i} \times d_{z,i}}$ (Section III.A)
**Part 2: Efficient Balanced Sampling Method for Extremely Sparse Tensors**
    Obtain $\mathcal{D} \in \mathbb{R}^{4 \times M_r \times n}$ from $\mathcal{B}$ utilizing **Algorithm 1**
**Part 3: Streaming Nonlinear Bayesian Tensor Decomposition (SNBTD) via Streaming patch processing**
    During training: Obtain low dimensional SNBTD embeddings
    $\mathcal{U}_{SNBTD}$ (Section III.C) to serve as a regularizer for Part 4
**Part 4: VAE with ANTLER loss**
    $(\phi, \theta, \theta_r) \leftarrow$ Initialize parameters
    **while** *SGD not converged* **do**
        $\mathcal{M} \sim \mathcal{D}$ (Random minibatch of data)
        $\epsilon \sim p(\epsilon)$ (Random noise for every data point in $\mathcal{M}$ (Reparameterization trick))
        Compute $\mathcal{L}_{\phi,\theta,\theta_r}(\mathcal{M}, \epsilon)$ from Eq. 5 and its gradients
        $\nabla_{\phi,\theta,\theta_r} \mathcal{L}_{\phi,\theta,\theta_r}(\mathcal{M}, \epsilon)$
        Update $\phi, \theta$ and $\theta_r$ using SGD optimizer
    **end**

---

An important property of the proposed loss function is the fact that it allows the joint optimization with respect to all parameters ($\phi$, $\theta$ and $\theta_r$) using stochastic gradient descent (SGD) algorithm. The parameters $\phi$, $\theta$ and $\theta_r$ can be initialized with random values and stochastically optimized until its convergence. Given the data is independent and identical distributed, the ANTLER objective is the sum (or average) of individual-datapoint ANTLER's, where SGD can be applied on the subset of data at a time:

$$\mathcal{L}_{\phi,\theta,\theta_r}(\mathcal{D}) = \sum_{\mathbf{x} \in \mathcal{D}} \mathcal{L}_{\phi,\theta,\theta_r}(\mathbf{x}) \quad (6)$$

One challenge of computing the gradient in the SGD algorithm is that the individual-datapoint ANTLER, and its gradient $\nabla_{\phi,\theta,\theta_r} \mathcal{L}_{\phi,\theta,\theta_r}(\mathbf{x})$, is intractable in general. However, for continuous latent variables and a differentiable encoder and generative model, the ANTLER loss can be differentiated with respect to $\phi$, $\theta$ and $\theta_r$ through a change of variables, also called the reparameterization trick [25, 26]. More specifically, the gradients $\nabla_{\phi,\theta,\theta_r} \mathcal{L}_{\phi,\theta,\theta_r}(\mathcal{M}, \epsilon)$ can be efficiently and automatically calculated via backpropagation. This procedure uses the information during the forward pass of model training to compute the gradient in the backward pass by performing the gradient operations in reverse order [27]. Therefore, we can still perform SGD with automatic differentiation.

### F. Tuning parameter selection

The ANTLER framework relies on three tuning parameters $(\lambda_1, \lambda_2, \lambda_3)$ in the loss function. $\lambda_1$ corresponds to the KL-divergence loss in the original ELBO loss function, which serves as a regularizer to the low dimensional VAE embeddings. This term which the representation space of $z$ to be meaningful and avoids overfitting. $\lambda_2$ corresponds to the term regularizing the VAE embeddings with the SNBTD embeddings. The benefit of this procedure is to take advantage of the Bayesian nonlinear, nonparametric tensor decomposition methods on small datasets and avoid the drawbacks of VAE embeddings mentioned in detail in Section III.E. Finally, $\lambda_3$ corresponds to the regression loss (Section III.D), which is the ultimate goal of the ANTLER framework. Therefore, this tuning parameter will have the highest weight and will be the objective function to be optimized during the tuning parameter selection procedure described in the remainder of this section. We note that the use of machine learning algorithms commonly involves careful tuning of learning parameters requiring expert experience, rules of thumb, or brute force search. On the contrary, we view this issue as the global derivative-free optimization of an unknown (nonconvex) black-box function and utilize the Bayesian optimization procedure proposed by [28] with its accompanying Python package "Spearmint" to automatically optimize the performance of the ANTLER algorithm for a given problem. Bayesian optimization has been shown to outperform other global optimization algorithms for tuning parameter selection on several multimodal black-box functions [29]. For further details on the Bayesian optimization, readers are referred to Snoek et al. [28] and the references therein.

## IV. SIMULATION STUDIES

In this section, we evaluate the ANTLER approach with simulated varying-size unstructured 3D point clouds against three benchmark methods. The common data characteristics are as follows: (i) the parts are measured and represented by unstructured point clouds of varying size, and (ii) the goal is to develop a model to link the point cloud inputs with a scalar or vector response. In the simulation studies, we will use two guiding examples to illustrate these characteristics: In the example of waveform surfaces the surface roughness serves as a response. For the conic shapes we predict the roundness error. Following Yan et al. [4], we consider a wave-shaped surface and truncated cones to generate point cloud objects. We simulated $N$ structured point clouds as training samples $X_i, i = 1, \ldots, N$ with $M$ measurement points. The simulated data are generated according to $X_i = M + V_i + E_i$ or in tensor notation as $\mathcal{X} = \mathcal{M} + \mathcal{V} + \mathcal{E}$, where $\mathcal{X}$ is a tensor combining the $X_i$'s , $\mathcal{M}$ is the mean of the point cloud data, $\mathcal{V}$ is the variational pattern of the point clouds, and $\mathcal{E}$ is a tensor of random noises. The noise is generated as white (iid) noise with $E_i \sim N(0, \delta^2)$. To simulate unstructured point clouds $X_i^u$ and generate a response with a varying number of measurements points, we apply the following procedure: 1) random subsampling is used to select $m_i$ out of $M$ points, where $m_i$ is determined by a uniform distribution (i.e. $m_i \sim U[m_l, m_u]$) with a respective lower bound ($m_l$) and an upper bound ($m_u$) on the number of measurement points. 2) Furthermore, we obtain three sorted lists of the 3D point clouds according to each of the three coordinate values individually and select the $\frac{m_r}{6}$ points with the smallest and the largest values respectively from each of the lists to obtain $m_r < m_u$ points in total. This will lead to an "extreme value" subset point cloud denoted by $X_i^{S,u}$ and result in areas with different point densities. 3) The scalar regression response $Y_i$ is computed by applying a nonlinear function $f$ on the unstructured point clouds $X_i^{S,u}$ of varying size (i.e., $Y_i = f_r(X_i^{S,u})$).

### A. Wave-shape Surface Point Cloud Simulation

Surface point clouds have important applications in surface prediction for precision manufacturing. Therefore, we simulate surface point clouds in a 3D Cartesian $(x, y, z)$ coordinate system, where $0 \leq x, y \leq 1$. The corresponding $z_{i_1, i_2}$ values at $\left(\frac{i_1}{I_1}, \frac{i_2}{I_2}\right)$ ($i_1 = 1, \ldots, I_1; i_2 = 1, \ldots, I_2$ with $I_1 = I_2 = 1,000$) for the $i$-th sample are stored in the matrix $X_i$, which is generated by $\mathcal{X} = \mathcal{V} + \mathcal{E}$. The variation patterns of the point cloud surface $\mathcal{V}$ are generated according to $\mathcal{V} = \mathcal{B} \times_1 U^{(1)} \times_2 U^{(2)} \times_3 Z$, where two mode-3 slices of $\mathcal{B} \in \mathbb{R}^{3 \times 3 \times 2}$ are generated as $B_1 = \begin{pmatrix} 4 & 1 & 0 \\ 1 & 0.1 & 0 \\ 1 & 0 & 1 \end{pmatrix}$ and $B_2 = \begin{pmatrix} 1 & 2 & 0 \\ 1 & 3 & 0 \\ 1 & 0 & 0.2 \end{pmatrix}$. Furthermore, three basis matrices $U^{(k)} = [u_1^{(k)}, u_2^{(k)}, u_3^{(k)}]$ are selected with $u_\alpha^{(k)} = [\sin\left(\frac{\pi\alpha}{n}\right), \sin\left(\frac{2\pi\alpha}{n}\right), \ldots, \sin\left(\frac{n\pi\alpha}{n}\right)]^T, \alpha = 1,2,3$. The input matrix $Z$ is sampled from a normal distribution $\mathcal{N}(0,1)$ [4]. Three examples of the generated surface point clouds are shown in Fig. 5.

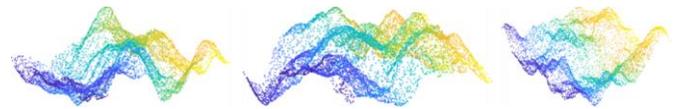

**Fig. 5**: Examples of generated wave-shape surfaces

An important application of 3D point clouds is the parameterization of surface roughness based on microwave remote sensing [30]. The obtained surface roughness is an important indicator in the modeling of surface dynamics, such as soil erosion or runoff estimation [30, 31].

Therefore, we compute the surface roughness $R_a$ based on point cloud data to serve as a regression response. First, a reference plane is defined as $z = \beta_0 + \beta_1 x + \beta_2 y$, where $x$, $y$ and $z$ are spatial coordinates in $\mathbb{R}^3$ and the vector of plane coefficients $\beta = [\beta_0, \beta_1, \beta_2]$ is obtained via orthogonal distance regression. In particular, $\beta = (X^T X - \delta^2 I)^{-1} X^T z$, where $X$ is the design matrix of $x$ and $y$ coordinates and $\delta$ is the smallest singular value of the augmented matrix $[X \ z]$. By scaling the normal vector $n$ to unit length, the orthogonal point-to-plane distance $D_j$ is computed as the positive inner product of the observation vector $v_i$ and the unit normal vector $n_u$ as $D_j = |n_u \cdot v_j|, j = 1, \ldots, m_r$. Finally, the surface roughness response is calculated as $R_{a,i} = Y_i = \sqrt{\frac{1}{m_r}\sum_{j=1}^{m_r}(D_j - \overline{D})^2}$ [32], which is a nonlinear function of the input point cloud $X_i^{S,u}$.

### B. Truncated Cone Point Cloud Simulation

In manufacturing applications, conic shapes are commonly used as a reference object for problems such as part-to-part variation pattern identification [33] or process control [4]. Therefore, following Yan et al.[4], we simulate truncated cone point clouds in a 3D cylindrical coordinate system $(r, \phi, z)$, where $\phi \in [0,2\pi]$ and $z \in [0,1]$. The corresponding $r$ values at $(\phi, z) = \left(\frac{2\pi i_1}{I_1}, \frac{i_2}{I_2}\right), i_1 = 1, \ldots, I_1, i_2 = 1, \ldots, I_2$ with $I_1 = I_2 = 100,000$ for the $i$-th sample are recorded in the matrix $X_i$. Furthermore, the variation patterns $\mathcal{V}$ of the point cloud surface are generated according to $r(\phi, z) = \frac{r + z\tan\theta}{\sqrt{1 - e^2\cos^2\phi}} + c(z^2 - z)$. Three different settings ($0.9\times, 1\times$, and $1.2\times$) of the normal conditions $\theta_0 = \frac{\pi}{8}$, $r_0 = 1.3$, $e_0 = 0.3$, $c_0 = 0.5$ corresponding to the different angles of the cone, radii of the upper circle, eccentricities of the top and bottom surfaces, and different curvatures, respectively. A full factorial design is used to generate training samples from different coefficient combinations. Four examples of the generated surface point clouds are shown in Fig. 6.

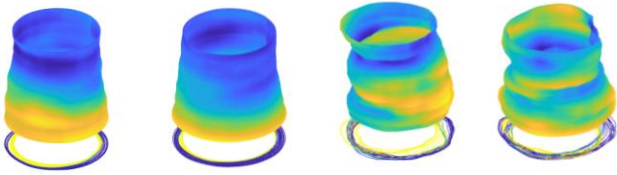

**Fig. 6**: Examples of generated truncated cones

An important quality characteristic of conic or cylindrical shape applications, such as hot steel rolling [34] or orthopaedical implants [35], is the roundness error of the manufactured parts. However, the application of the minimum zone tolerance (MZT) method, which provides the most accurate estimation of the roundness error, is computationally intensive [36]. However, the proposed ANTLER framework can directly estimate such quality characteristics based on a point cloud input. Therefore, we will utilize the MZT method to compute the quality response in this simulation study via a nonlinear function $f_r(\cdot)$ in the following way. Given a circumferential line $r(x, y, \phi)$, the roundness error $R(x, y)$ is defined by

$$R(x,y) = OC(x,y) - IC(x,y), \quad (x,y) \in E_{r(x,y,\phi)}, \quad (7)$$

where $OC(x,y)$ and $IC(x,y)$ are the radii of the reference circles of center $(x,y)$ derived by $n$ finite measurement points and $E_{r(x,y,\phi)}$ is the area enclosed by $r(x,y,\phi)$:

$$OC(x,y) = \max_{\phi_i = i\times\frac{2\pi}{n}, i=1,\ldots,n} r(x,y,\phi_i) \quad (8)$$

$$IC(x,y) = \min_{\phi_i = i\times\frac{2\pi}{n}, i=1,\ldots,n} r(x,y,\phi_i) \quad (9)$$

Finally, the average roundness $\overline{R}(x,y)$ over all circumferential lines is used as the response $Y_i$.

### C. Simulation Study Prediction Results

In this section, we will compare the proposed ANTLER framework with three existing methods.

From the field of tensor regressions, we will use the Multiple Tensor on Tensor (MTOT) regression [6] as a benchmark. To make these methods applicable for unstructured point cloud data, we will use the binary sampling tensor $\mathcal{D}$ obtained from the raw measurement point cloud in order to make the data structured and applicable to the MTOT method. We note, that this method heavily favors the MTOT method since it utilizes the efficient representation and Algorithm 1 of our proposed framework. The MTOT framework by itself could not handle the unstructured, varying-size 3D point cloud inputs directly.

As a deep learning benchmark, we will use a modified PointNet architecture [1]. This method was originally designed for classification tasks but can be modified for regression tasks by adjusting the output layers activation function from a normalized exponential (Softmax) to a Rectified Linear Unit (ReLU) function. In the remainder of this paper, this method is referred to as PointNet Regression (PointNetR).

Furthermore, we will use a classical machine learning method to evaluate our proposed algorithm. Previous research by Biehler [37] has compared multiple machine learning methods for regression tasks with unstructured point cloud input and concluded that Random Forest (RF) outperforms all other benchmarks. Therefore, we will only use the RF as a benchmark method in the simulation and case study. The features are extracted inspired by the "key to the PointNet approach" [1], which uses a symmetric function (e.g., maximum) to aggregate the information from point clouds. In particular, a min and max function was applied to obtain the 10,000 largest and smallest points of each coordinate axis. Several other ways of feature extraction such as covariance analysis of the distance-driven local neighborhoods [38], Riemann graphs over local neighborhoods[39], multiscale and hierarchical point clusters [40] and a self-organizing network for point cloud analysis [41] were also investigated but yielded inferior results.

For 3D measurement acquisition devices, the number of measurements points in different samples usually does not vary by order of magnitudes. Since we chose the lower bound of points as the number for subsampling, the resulting fixed size point clouds usually contain at least half of the original measurement points. Therefore, the choice of subsampling technique does not significantly influence the results. However, more advanced subsampling techniques, such as important support points [42] to find an "optimal" subsample from weighted proposal samples, could be applied under extreme measurement conditions. In each case, we compare the proposed method with benchmark methods based on the Root Means Squared Error (RMSE) calculated at different levels of noise $\delta$. Table I reports the average and standard deviation of RMSE obtained via 10-fold Cross Validation for the simulation cases 1 (Waveform Surface) and 2 (Truncated Cone), respectively.

## TABLE I
RMSEs AND STANDARD DEVIATION (SD) FOR CASE 1 AND 2
(NOTE: BOLD INDICATES THE SMALLEST VALUE IN EACH ROW COMPARED TO ITS CORRESPONDING VALUES IN OTHER ROWS.)

| Method | Case 1: Waveform Surface | | Method | Case 2: Truncated Cone | |
|---|---|---|---|---|---|
| $\delta = 0.1$ N=100 | RMSE | SD | $\delta = 0.01$ N=100 | RMSE | SD |
| MTOT | 18.832 | 4.687 | MTOT | 31.784 | 9.273 |
| PointNetR | 16.938 | 4.899 | PointNetR | 32.047 | 10.237 |
| RF | 11.442 | 3.107 | RF | 14.367 | 3.909 |
| ANTLER | **6.768** | **2.424** | ANTLER | **9.874** | **3.782** |
| $\delta = 1$ N=100 | RMSE | STD | $\delta = 0.1$ N=100 | RMSE | STD |
| MTOT | 38.048 | 10.102 | MTOT | 43.929 | 6.289 |
| PointNetR | 50.444 | 46.620 | PointNetR | 47.406 | 8.190 |
| RF | 34.109 | 7.855 | RF | 34.883 | 6.328 |
| ANTLER | **14.297** | **4.842** | ANTLER | **21.372** | **5.081** |

In all cases, the proposed ANTLER framework outperforms the benchmark methods MTOT, PointNetR, and RF with the smallest RMSE error, reflecting the advantage of the ANTLER method in terms of prediction accuracy. The performance improvement is mainly due to the following three reasons: (i) an efficient tensor representation of the point clouds, (ii) a nonlinear Bayesian tensor decomposition, and (iii) nonlinear regression models integrated by a unified VAE framework. Furthermore, with an increase in $\delta$, all methods exhibit a larger RMSE and variance of the prediction results.

The other benchmark methods cannot outperform the proposed ANTLER due to their own limitations as follows: (i) Even with significant help from our proposed point cloud representation and Algorithm 1, MTOT method cannot handle the unstructured nature of the point clouds effectively and relies on the linear assumption. Therefore, it fails to capture the complex, nonlinear relationships between unstructured point clouds and a scalar response. (ii) The modified PointNet also performs worse since it fails to efficiently represent the high dimensional input data as a low dimension manifold for small sample sizes. Due to the deep learning architecture, a very large number of parameters needs to be estimated from a limited number of samples leading to overfitting. (iii) The Random Forest does not exploit the rich spatial information contained in point clouds by using a two-step approach, which does not consider the regression response during feature extraction. Additionally, the features extraction fails to capture small local features highly correlated with the regression response.

## V. CASE STUDY

In this section, a real case study concerning gears manufactured by micro gear hobbing is introduced and used as a guiding example for the use of the proposed ANTLER methodology. Due to the high engineering tolerance requirements, numerous manufacturing defects occur, causing increased costs. However, in the current state of the art, the quality of the final product, such as gear boxes, can only be ensured through an end-of-line test bench or extensive FEA simulations [43]. However, predicting functional characteristics of gears during cycle time is the cornerstone of selective assembly and the optimization of gear noise emissions. This is of particular interest in applications such as automotive or medical engineering.

To advance the state of the art and achieve in-process quality improvements (IPQI), the gears need to be measured during the cycle time. Subsequently, based on the obtained point clouds, the functional characteristics of interest need to be predicted. By identifying defective parts in terms of their engineering tolerance and function fulfillment, a more holistic approach toward the identification of scrap and selective assembly can be achieved.

### A. Gear Dataset

The case study is based on 120 optically measured gears as described in preliminary work by Gauder et al.[44, 45]. The gears were measured in a climate-controlled measuring room on an Alicona µCMM. The µCMM is a highly accurate, purely optical, non-contact 3D micro-coordinate measurement machine and enables the measurement of dimensions, position, shape, and surface roughness. The results of the measurement are three-dimensional point clouds, as shown in Fig. 7.

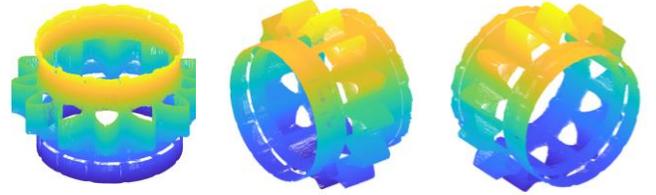

**Fig. 7**: Gear measurement point clouds

Due to the manual clamping, the orientation and rotation of the measurement point clouds may not be aligned. To align the point clouds, the CAD model is used as a reference, and the rotation and translation matrix of the point clouds is obtained via the Iterative Closest Point (ICP) algorithm [46]. After the initial experiments, where the clamping process was conducted manually, the clamping of the gears in the measurement device will be automated, and the process of alignment via ICP can be skipped. To obtain the transmission error and eccentricity in both directions of rotation, the gear-specific FEA software ZaKo3D is utilized [47]. The input data to the simulation are geometric measurement data (i.e., point clouds). Based on the gear measurements, a finite element structure is generated, and by applying a nonlinear time-varying spring model, several outputs such as transmission error, loads, and deflects on the tooth are simulated. The FEA computation time for one sample is approximately 2 hours, so much longer than the cycle time of approximately 3 minutes For further background on the nonlinearity of the simulation outputs, readers are referred to [48, 49]. In the remainder of this paper, the four response variables for the regression framework are the following: the clockwise transmission error (TE) and eccentricity (E) are denoted by $Y_1$ and $Y_2$, while the counterclockwise TE and E are denoted by $Y_3$ and $Y_4$, respectively.

During the processing of the raw point clouds, the origin and z-axis orientation are assumed to be fixed for each point cloud. For the alignment of point clouds, we propose two strategies depending on the measurement setup: In controlled environments, where the measurement objects can be approximately aligned before each measurement, the point clouds can be used directly. If no alignment of subsequent samples can be guaranteed, the ICP method can be adopted for alignment of the point clouds.

### B. Case Study Prediction Results

To develop a predictive model, perform 10-fold cross validation (CV). In this paper, we compared the same three benchmark methods as the simulation study, random forest (RF), Multiple Tensor-on-tensor regression (MTOT), and a deep learning method PointNet regression (PointNetR), which are applied on the aligned datasets. Fig. 8 depicts the boxplot of root mean squared errors (RMSE) obtained over 120 measurements for the proposed method and the benchmarks.

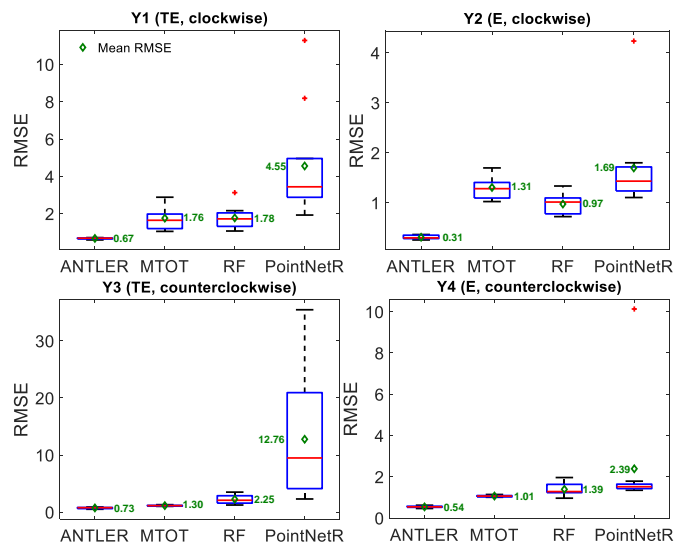

**Fig. 8**: Boxplot of RMSE's for predicting the functional response using ANTLER, MTOT, PointNetR and RF

The results show the superiority of the proposed ANTLER framework in comparison to three benchmark methods. On average, across all CV scores and response variables, the RSME of the ANTLER method is 258.5%, 285.6%, and 850.4% smaller than the respective benchmark methods MTOT, PointNetR, and RF, respectively. In particular, the adapted PointNet, which is a state-of-the-art method in point cloud modeling, exhibits a high variance due to the small sample sizes and complex, nonlinear structure of the learning task. ANTLER predicts the scalar response more accurately, capturing the complex nonlinear relationship by exploiting the information in unstructured point clouds of varying size in a simultaneous and disciplined manner. The results make a case for point cloud modeling in complex, nonlinear regression settings.

## VI. CONCLUSION

Discriminative models for unstructured point cloud data are an emerging research area, with applications in smart manufacturing, geological surveying, or autonomous driving due to the capabilities of 3D scanners in accurate shape modeling. To address an important industry problem and several research gaps in the processing of unstructured, varying-size point clouds, we proposed a novel ANTLER framework. This approach directly uses the entire information contained in the high-dimensional unstructured point cloud with varying sizes, structures it as a binary tensor, utilizes a balanced sampling strategy and further employs efficient algorithms for simultaneous tensor decomposition and nonlinear regression. Two simulation studies and a real-world case study illustrate the effectiveness of the proposed approach for real-time in-process functional prediction based on an unstructured 3D point cloud input of varying sizes. Future work could extend this framework to the detection and classification of multiple types of surface anomalies. Moreover, the application of the high-level idea to other tensor learning problems (e.g., nonlinear multiple tensors on tensor regression) is a promising direction. Finally, tensor completion or robust tensor estimation techniques could be applied to explicitly handle missing values or noise in the input point clouds.

## Acknowledgment

This research is partially funded by the NSF CNS 2019378 and NSF CMMI 1922739.

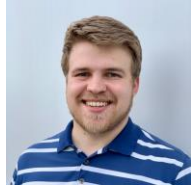
**Michael Biehler** (Member, IEEE) received his B.S. and M.S in Industrial Engineering and Management with a major in production engineering from Karlsruhe Institute of Technology (KIT) in 2017 and 2020, respectively. Currently, he is a Ph.D. student in the Stewart School of Industrial and Systems Engineering, Georgia Institute of Technology. His research rests at the interface between machine learning and cyber physical (manufacturing) systems, where he aims to develop methods for cyber security, monitoring, prognostics, and control. He is a member of ASME, ENBIS, IEEE, IISE, INFORMS and SIAM.

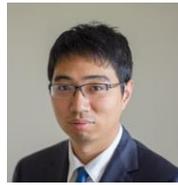
**Hao Yan** received the B.S. degree in physics from Peking University, Beijing, China, in 2011, and the M.S. degree in statistics, the M.S. degree in computational science and engineering, and the Ph.D. degree in industrial engineering from the Georgia Institute of Technology, Atlanta, GA, USA, in 2015, 2016, and 2017, respectively. He is currently an Assistant Professor with the School of Computing, Informatics, and Decision Systems Engineering (SCIDSE), Arizona State University (ASU), Tempe, AZ, USA. His research interests focus on developing scalable statistical learning algorithms for large-scale high-dimensional data with complex heterogeneous structures to extract useful information for the purpose of system performance assessment, anomaly detection, intelligent sampling, and decision making. Dr. Yan was a recipient of multiple awards including the Best Paper Award in the IEEE Transactions on Automation Science and Engineering and the ASQ Brumbaugh Award. He is a member of INFORMS and IIE.

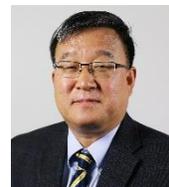
**Dr. Jianjun Shi** received the B.S. and M.S. degrees in automation from the Beijing Institute of Technology in 1984 and 1987, respectively, and the Ph.D. degree in mechanical engineering from the University of Michigan in 1992. Currently, Dr. Shi is the Carolyn J. Stewart Chair and Professor at the Stewart School of Industrial and Systems Engineering, Georgia Institute of Technology. His research interests include the fusion of advanced statistical and domain knowledge to develop methodologies for modeling, monitoring, diagnosis, and control for complex manufacturing systems. Dr. Shi is a Fellow of four professional societies, including ASME, IISE, INFORMS, and SME, an elected member of the International Statistics Institute (ISI), a life member of ASA, an Academician of the International Academy for Quality (IAQ), and a member of National Academy of Engineers (NAE).